\documentclass[a4paper, 11pt]{article}

\usepackage{a4wide}
\usepackage{amsmath}
\usepackage{amssymb}
\usepackage{commath}
\usepackage{latexsym}
\usepackage{graphicx}
\usepackage{color}
\usepackage{verbatim}
\usepackage{authblk}
\usepackage{enumitem}
\usepackage{nicematrix}
\usepackage{algorithm}
\usepackage{algpseudocode}
\usepackage{booktabs}
\usepackage{tabularx}
\usepackage[pdfstartview=FitH]{hyperref}

\setlist[enumerate,1]{label=(\arabic*), ref=\arabic*}
\setlist[enumerate,2]{label=(\arabic{enumi}\alph{enumii}), ref=\arabic{enumi}\alph{enumii}}

\newenvironment{keywords}
{\small\noindent\textit{Keywords: }\ignorespaces}
{\par\medskip}

\title{Robustness of neural networks to random \\ noise perturbations of their inputs}

\author{Mark Levene and Martyn Harris \\
School of Computing and Mathematical Sciences, \\
Birkbeck, University of London\\
Malet Street, London WC1E 7HX, U.K.\\
\texttt{m.levene@bbk.ac.uk,m.harris@bbk.ac.uk}}

\date{}

\begin{document}

\maketitle

\begin{abstract}

We investigate the problem of the robustness of a trained neural network to the perturbation of its input values. More specifically, we examine the interplay between the accuracy of the network, as measured by the mean squared error, and robustness. Accordingly, we present a robustness measure, which, with high probability, suggests an upper bound on the mean squared error of the network, with respect to an input data set, for a given perturbation of the input values of the network. The measure we propose is both simple and efficient to compute, treating the neural network as a black box. We provide experimental results on several real-world data sets showing the efficacy of the proposed method. We also introduce the concept of robustness curves, which allows us to further analyse robustness within and between data sets. 

\end{abstract}


\section{Introduction}

Data science along with machine learning (ML) \cite{LEVE25}, and especially {\em neural networks} (NN) \cite{AGGA18} have been in the forefront of the recent resurgence in artificial intelligence (AI) both in academia and in industry. The evaluation of an AI system is the measure of its success, that is, for NNs, determining the quality of the trained model. 

\smallskip

In particular, the {\em accuracy} of a NN is measured by the proportion of correct predictions it makes on the test data. In the following we will use the term {\em accuracy} as an umbrella for evaluation measures in ML \cite{JAPK11} and, in particular, for NNs.
For regression problems, on which we mainly focus in this work, measures such as {\em mean squared error} (MSE) and {\em mean absolute error} (MAE) are commonly used \cite{LEVE25}; here we will concentrate on MSE. (For classification problems we use a variant of the MSE, called the {\em multi-class Brier score} (MBS) \cite[Section 4.4]{SCRU23}, which is the counterpart of the MSE that can be used for classification problems see \cite{HUI21}; unless otherwise specified, in the following, MSE refers to both MSE and MBS.)

\smallskip

Moreover, we also require that the predictions are stable under a wide range of conditions such as noise in the input, which may be caused by random or adversarial perturbations of the input. This stability of an ML model such as an NN is called {\em robustness} \cite{ALIP09,SZEG14,MADR18,BALE25}. There are trade-offs between accuracy and robustness, where the increase of one may cause a decrease of the other; these trade-offs are reviewed in \cite{LI25}. Here we concentrate on the robustness to input perturbations resulting from adding random Gaussian noise to the original input \cite{RAVI99,ZUR09}. 
We note that although we focus on NNs, the method we propose is valid for other ML algorithms, since, for the purpose of this paper, we treat a NN as a black box.

\smallskip

Our notion of robustness of a NN is both intuitive, relatively simple, and can be computed in a straightforward manner. It hinges on adapting the MSE for perturbations with a fixed standard deviation, and then computing the robustness of the NN using a Monte Carlo method \cite{ROSS23} to obtain a confidence interval for the derived MSE. This contrasts with other research on robustness, such as \cite{ALIP09}, which focuses on perturbation of network weights by modifying the model, and is limited to classification problems \cite{SZEG14,MADR18}.

\smallskip

Apart from the standard evaluation on real-world data sets, with accuracy and MSE, we also introduce {\em robustness curves}, which visualise in a clear manner how increasing the perturbation of the input will degrade the MSE (see \cite{SIRC21}). Then, in order to model robustness curves, we fit the curves to the {\em Gompertz function} \cite{SEBE03,TJOR17}, which is widely used for growth processes, saturation effects, and cumulative phenomena. This allows us to characterise the growth rate of the curves, identify a linear regime in which the MSE grows slowly, and to compare robustness across data sets.

\smallskip

The rest of the paper is organised as follows.
In Section~\ref{sec:methods} we present the methods used, including a brief review of the background on NNs needed herein, the method of perturbation by adding random Gaussian noise to the test data, and the evaluation of such perturbations. 
In Section~\ref{sec:robust} we formalise a measure of the robustness of a NN, assuming that the NN has already been trained and its inputs are perturbed with Gaussian noise.
In Section~\ref{sec:exp} we describe experiments we carried out on real-world data sets to validate our approach, and in Section~\ref{sec:eval} we present a detailed evaluation of the experimental results, including the novel use of robustness curves to enhance the analysis and compare the performance of robustness within and between the employed data sets.
Finally, in Section~\ref{sec:conc}, we present our concluding remarks, and in Appendix~\ref{app:gompertz} we provide additional figures and tables. 

\section{Methods}
\label{sec:methods}

In Subsection~\ref{subsec:nn} we briefly review NNs and introduce the notation adopted in this paper. In Subsection~\ref{subsec:perturb} we introduce perturbation of the inputs, and how the definition of the MSE can be adapted to deal with these perturbations.

\subsection{Neural networks}
\label{subsec:nn}

\begin{figure}[htp]
\begin{center}
\includegraphics[width=12cm]{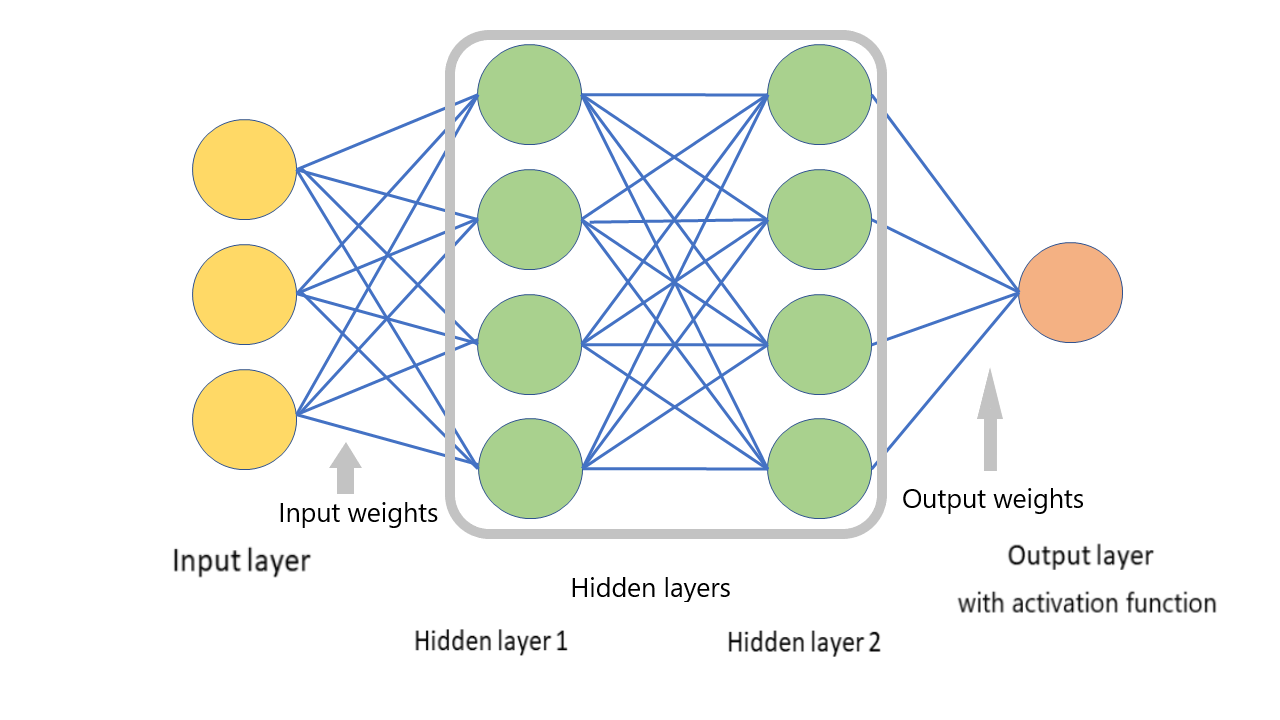}
\caption{\label{fig:nn} A feedforward neural network with two hidden layers.}
\end{center}
\end{figure}

To simplify the model under discussion, let us adopt a generic neural network with input and output layers and some hidden layers inbetween, such as the {\em feedforward neural network} (FNN) shown in Figure~\ref{fig:nn}; the hidden layers may be viewed as the latent space of the network. In general, our methods may be deployed at various parts of the neural network architecture, although here will concentrate on the effect that perturbing an input will have on the predicted output of the network. Moreover, the methods we present are intended to be generic and, in principle apply to both regression or classification problems, however, since we are focusing on input perturbation method we will assume, for the presentation of the method, that the network's inputs are continuous.

\smallskip

We assume that a neural network model, say NN, has been trained, and we have a test data set (or simply a data set) $T$ containing a sample of $n$ pairs (i.e. its cardinality is $n$, written as $|T| = n$). Each pair, $(x_i, y_i)$ in $T$ is such that $x_i$ is a feature vector of fixed dimension (often called the {\em input value}), with $|x_i| \ge 1$ (called the {\em width of the input value}), and $y_i$ is also a feature vector of fixed dimension  (often called the {\em actual value}) typically satisfying $|y_i| = 1$.  Thus the data set $T$ induces a finite input set, $x = \{x_1,x_2, \ldots, x_n\}$, with corresponding target output values (or simply target values), $y = \{y_1,y_2,\ldots,y_n\}$. The set of predicted output values (or simply predicted values) resulting from running the neural network model NN on the input set $x$ is given by $\hat{y} = \{\hat{y}_1, \hat{y}_2, \ldots, \hat{y}_n\}$. 

\smallskip

We do not specify the train and test method of the neural network, which, for example, could be cross-validation \cite{AGGA18}; we evaluate the trained NN using {\em mean squared error} (MSE) and make use of the {\em root mean squared error} (RMSE) which has the same units as the target values \cite{CHIC21a,LEVE25}. (We note that there is also an argument for using the MAE, a measure known to be more robust to outliers \cite{CHAI14}, however we leave this choice for future work.)

\smallskip

Now, given that the NN models the function computed by the neural network, we have
\begin{equation}\label{eq:nn}
\hat{y}_i = {\rm NN}(x_i),
\end{equation}
where $x_i$ is its input value and $\hat{y}_i$ is the predicted output value.

\subsection{Perturbing the inputs}
\label{subsec:perturb}

For simplicity we will assume that, $|x_i| = 1$, that is, each input value is a single feature, that the variance of $x_i$ is $1$, and as stated in Subsection~\ref{subsec:nn}, $x_i$ is continuous. These assumptions will be relaxed in Section~\ref{sec:exp} when we experiment with real-world data sets that do not satisfy these assumptions. 

\smallskip

We perturb the input with randomly generated Gaussian noise, noting that in our case the model is fixed and thus noise is injected into the input during testing rather than during training as in \cite{RAVI99,ZUR09}, where the objective is to improve the performance of the neural network.

\smallskip

Now, assume that $\epsilon$ is a normally distributed random variable, with mean $0$ and variance $\sigma^2$, i.e.\ from ${\cal N}(0,\sigma^2)$, and that $\sigma$ is a small non-negative value; typically $\sigma < 1$ but ultimately its value will also depend on the data set. In the special case when $\sigma = 0$ we have the degenerate case of $\epsilon = 0$. (For simplicity, we will use the symbol $\epsilon$ to denote both the random variable and its realised value, with the intended meaning being clear from context.)

\smallskip

As stated in Subsection~\ref{subsec:nn}, we choose to make use of the MSE as our loss function. Thus, {\em mean squared error} (${\rm MSE}\epsilon$) of NN with respect to $T$ and perturbation $\epsilon$, is given by
\begin{equation}\label{eq:mse}
{\rm MSE}\epsilon({\rm NN}) = \frac{1}{n} \ \sum_{i=1}^n \left( y_i - {\rm NN}(x_i + \epsilon) \right)^2,
\end{equation}
bearing in mind that for $\epsilon = 0$ and we have the standard MSE (which we also refer to as the {\em baseline MSE}), where $y_i$ is the target output and ${\rm NN}(x_i) = \hat{y_i}$ is the predicted output as in (\ref{eq:nn}).

\section{A measure of robustness}
\label{sec:robust}

We introduce a novel measure of robustness for NNs, called $\theta$-{\em robustness}, and show how the MSE can be redefined to make use of this measure, leading to ${\rm MSE}\theta$.

\smallskip

Now, a neural network NN is said to be $\theta$-{\em robust} with significance $\alpha$, $0 < \alpha \le 1$, for standard deviation $\sigma$, with $\rho$ being a function of $\sigma, \alpha$ and NN, defined as
\begin{equation}\label{eq:robust-mse}
\rho(\sigma, \alpha, {\rm NN}) = \min_\theta \ {\rm such} \ {\rm that} \ P \! \left({\rm MSE}\epsilon({\rm NN}) \le \theta \right) \ge 1-\alpha,
\end{equation}
where $\theta \ge 0$ and $\epsilon$ is a random variable distributed as ${\cal N}(0,\sigma^2)$. 

\smallskip

To recap, given a trained neural network, NN, and $\sigma$, (\ref{eq:robust-mse}) gives us an upper bound on ${\rm MSE}\epsilon({\rm NN})$, given in (\ref{eq:mse}), at the $\alpha$ confidence level. It is important to note that smaller values of $\theta = \rho(\sigma, \alpha, {\rm NN})$ correspond to higher levels of robustness.
Pseudocode for computing an approximation of $\rho(\sigma, \alpha, {\rm NN})$ is given in Algorithm~\ref{alg:robust}. (See \cite{IALO19} detailing other non-parametric methods for estimating confidence intervals for quantiles.) 

\algrenewcommand{\algorithmiccomment}[1]{\hfill\% #1}
\renewcommand{\algorithmicrequire}{\textbf{Input:}}
\renewcommand{\algorithmicensure}{\textbf{Output:}}
\begin{algorithm}[ht]
\caption{Robustness-computation}\label{alg:robust}
\begin{algorithmic}[1]
\Require $\sigma, \alpha, {\rm NN}, m$
\Ensure $\theta$
\For {$j=1:m$} \Comment $m=1000$ is reasonable
\State Generate $n$ random values, $\epsilon_i$, each distributed as ${\cal N}(0,\sigma^2)$
\State $\theta_j = {\rm MSE}\epsilon({\rm NN})$ using the generated $\epsilon_i$ values
\EndFor
\State Arrange the $\theta_j$ values in ascending order to obtain a sequence $\theta_{(1)}, \theta_{(2)} \ldots, \theta_{(m)}$
\State $ub = \left\lceil (1-\alpha) m \right\rceil$ \Comment For convenience we may choose $m$ such  that $\left\lceil (1-\alpha) m \right\rceil  = (1-\alpha) m$
\State $\theta = \theta_{(ub)}$
\State \Return ($\theta$)
\end{algorithmic}
\end{algorithm}
\smallskip

We mention a similar definition of robustness in \cite{ALIP09}, who are concerned with perturbations of the NN weights over a perturbation space, which can be sampled from a provided probability distribution function over the space. In particular, in their experiments, \cite{ALIP09} considered the uniform distribution for sampling the perturbation space. Moreover, the exact nature of the perturbation in the theoretical model is not specified in \cite{ALIP09}. In contrast, our model does not depend on the architecture of the NN as such as long as the NN is deterministic once trained. In our model we investigate the robustness of a fixed NN, defined with respect to perturbations of the inputs induced by ${\cal N}(0,\sigma^2)$, and investigate how the robustness changes with $\sigma$.

\smallskip 

We observe that ${\rm MSE}\epsilon$ can be viewed as a measure of robustness under perturbations, governed by $\sigma$. In addition, when $\sigma_1 \le \sigma_2$, we expect $\rho(\sigma_1, \alpha, {\rm NN}) \le \rho(\sigma_2, \alpha, {\rm NN})$. Thus, in general, we expect ${\rm MSE}_0$ (i.e. \!the baseline MSE) to give rise to a lower bound baseline for $\theta$, although for small $\sigma$, ${\rm MSE}\epsilon$ may well fluctuate around ${\rm MSE}_0$. 

\smallskip

We further note that it is obviously beneficial for ${\rm MSE}_0$ to be as small as possible without overfitting the training data, since if the model does not fit the data well, then measuring robustness does not make sense. (See \cite{LI25} for a discussion on the tradeoff between accuracy measures, such as mean squared error, and robustness.)

\smallskip

We can now define the mean squared error of the difference between the predicted output of the neural network on an input, $x_i$, and its predicted output on a perturbed input, $x_i + \epsilon$, as,
\begin{equation}\label{eq:robust-theta}
{\rm MSE}\theta({\rm NN}) = \frac{1}{n} \ \sum_{i=1}^n \left( {\rm NN}(x_i) - {\rm NN}(x_i + \epsilon) \right)^2,
\end{equation}
where NN is $\theta$-robust and  $\epsilon$ is distributed as ${\cal N}(0,\sigma^2)$ with $\theta = \rho(\sigma, \alpha, {\rm NN})$, noting that $\theta$ depends on $\sigma$, which in turn derives $\epsilon$. 

\smallskip

We mention that there is a growing body of research looking into conditions that guarantee that the difference between ${\rm NN}(x_i + \epsilon)$ and ${\rm NN}(x_i)$ is bounded for small $\epsilon$. In particular, this may be theoretically achieved by employing {\em Lipschitz continuity} for the function represented by the neural network; see \cite{ZUHL25}. However, there are some limitations of this approach in terms of expressiveness and computation efficiency \cite{HUST19}. Although we do not discount the potential benefits of the Lipschitz continuity approach, here we restrict ourselves to measuring robustness from an experimental data-driven perspective based on (\ref{eq:robust-mse}).

\section{Experimental setup}
\label{sec:exp}

In Subsection~\ref{subsec:ds} we describe the test data sets used to assess our notion of robustness with respect to the neural networks trained on their respective training data sets, while in Subsection~\ref{subsec:real} we relax the assumptions presented in Subsection~\ref{subsec:perturb} regarding the perturbation of the data for different types of features. 
Our objective, is then, in Section~\ref{sec:eval}, to determine how robust the neural network is, that is, determine the minimal $\theta$ according to (\ref{eq:robust-mse}), which allows the computation of ${\rm MSE}\theta$ according to (\ref{eq:robust-theta}). 
    
\subsection{Data sets}
\label{subsec:ds}

In Table~\ref{tbl:datasets} we describe the chosen real-world data sets. In particular, we summarise key attributes of each data set, including the number and types of input features, noting that there is only one target feature.

\begin{table}[htbp]
\centering
\small
\renewcommand{\arraystretch}{1.0}
\begin{tabularx}{\textwidth}{X c c X X}
\toprule
Data set & Task & $n$ & Input type & Target type \\
\hline
Deaths from Cancer \cite{OECD_deaths} &
regression &
31 &
continuous\:(1): single temporal variable (calendar year). &
continuous: Cancer mortality rate; deaths per 100,000 population. \\
\hline
Flu vaccination rates in 2020 \cite{OECD_flu} &
regression &
31 &
continuous\:(1): ordered temporal index. &
continuous: Flu vaccination uptake rate; percentage. \\
\hline
Cleveland, Heart disease \cite{UCHD26} &
classification &
237 &
continuous\:(6), ordinal\:(3), categorical/nominal\:(19): encoded clinical attributes from clinical measurements and diagnostic codes. &
Binary disease indicator. \\
\hline
Wisconsin diagnostic, Breast Cancer  \cite{WDBC26} &
classification &
569 &
continuous\:(30): geometric and textural features from cell nuclei images. &
Binary disease indicator. \\
\hline
MNIST \cite{MNIS19} &
classification &
70,000 &
continuous\:(784): flattened $28 \times 28$ grayscale pixel intensities scaled to $[0,1]$. &
categorical digit class $\{0,\dots,9\}$. \\
\bottomrule
\end{tabularx}
\caption{Summary of data sets used in the robustness experiments, including the types and number of input features, noting that there is only one target feature.}
\label{tbl:datasets}
\end{table}

\subsection{Perturbing real data}
\label{subsec:real}

Here we relax the assumptions to the perturbation of attribute values as given in Subsection~\ref{subsec:perturb}. In particular, we only perturb numeric features and do not alter categorical features \cite{LEVE25}, for example as is proposed in \cite{PRAK24}. To make this decision more concrete, as an example, it does not make sense to perturb a categorical attribute such as gender, since changing this value will completely alter the meaning of the data item.

\smallskip

Now, let the feature vector $x_i$ have $m$ numeric features, $m \ge 1$, and thus 
\begin{displaymath}
x_{i1}, x_{i2}, \ldots, x_{im},
\end{displaymath}
are the $m$ numeric features of $x_i$, where the empirical standard deviation of $x_{ij}$ is $\sigma_{ij}$, with $1 \le j \le m$. (For simplicity we assume that the numeric features of $x_i$ precede the categorical ones.) So, the perturbation $\epsilon_{ij}$ that we apply to the $j$th numeric feature of $x_i$, will be a normally distributed random variable with mean $0$ and variance $(\sigma \sigma_{ij})^2$, i.e. from ${\cal N}(0,(\sigma \sigma_{ij})^2)$, that is, we scale ${\cal N}(0,\sigma^2)$ by $\sigma_{ij}^2$. 

\smallskip 

We observe that $x_i$ may have more than $m$ features, and so any additional categorical features are not perturbed. So, $\epsilon$ in (\ref{eq:mse}) is actually a vector of width $|x_i|$, where $m$ of the features are perturbed and the rest are not. Moreover, when the numeric value of the $j$th numeric feature is discrete rather than continuous, we round the perturbed value, that is, the corresponding ${ij}$ component in $x_i$ becomes $round(x_{ij} + \epsilon_{ij})$ rather than $(x_{ij} + \epsilon_{ij})$.

\smallskip

Whenever $m > 1$, we make use of a probability $p$, which is used to decide for each numeric data item of a feature, with probability $p$, whether to perturb its value or not; we call $p$ the {\em perturb or not probability}. In our experiments $p$ takes on four values $0.25,0.5,0.75$ and $1$, noting that when $m=1$ we perturb all input data items, i.e. $p=1$.

\section{Evaluation}
\label{sec:eval}

In Subsection~\ref{subsec:numerics} we describe and summarise the numeric results, while in Subsection~\ref{subsec:curves} we describe and summarise robustness curves, which aid our understanding of how the robustness results change when the standard deviation $\sigma$ used in perturbing the data values is increased.
 
\subsection{Summary of numeric results}
\label{subsec:numerics}

Prior to presenting the results we describe a variant of the MSE used for evaluating classification problems rather than regression ones. The metric we use for classification is called the {\em multi-class Brier score} (MBS) \cite[Section 4.4]{SCRU23}, which is the counterpart of the MSE for classification problems; see \cite{HUI21}. 

\smallskip

For $K > 1$ classes, the ${\rm MBS}\epsilon$  of NN with respect to the data set $T$ and perturbation $\epsilon$, is defined as
\begin{equation}\label{eq:brierx}
{\rm MBS}\epsilon = \frac{1}{2 n}\sum_{i=1}^n \sum_{k=1}^K \bigl(y_{ik} - p_{ik}\bigr)^2,
\end{equation}
where 
\begin{displaymath}
{\rm NN}(x_i + \epsilon) = (p_{i1}, p_{i2}, \dots,p_{iK})
\end{displaymath}
is the vector of the $K$ predicted class probabilities, with $x_i$ and $\epsilon$ being $|x_i|$ vectors as explained in Section~\ref{sec:exp}, and 
\begin{displaymath} 
y_i = (y_{i1}, y_{i2},\dots,y_{iK})
\end{displaymath}
being the one-hot encoding \cite{LEVE25} $K$-vector for the $i$-th observation, where $y_{ik} = 1$ if class $k$ is the target class, and $0$ otherwise, for $1 \le k \le K$. 
(From now on, for notational convenience, we will not distinguish between ${\rm MBS}_\epsilon$ and ${\rm MSE}_\epsilon$ and often refer to them both simply as ${\rm MSE}\epsilon$; ${\rm MBS}_0$, the baseline MBS, is referred to as the baseline MSE.)

\smallskip

The interpretation of  ${\rm MBS}_\epsilon$ for input perturbation of $\epsilon$, is thus the 
average squared Euclidean distance between the predicted probability vector and the one-hot encoding, rescaled so that a maximally wrong prediction scores 1; see \cite{HOES26}.

\smallskip

In Table~\ref{tbl:arch} we show for each data set in the experiments the FNN architecture used and how the data was split, including both the depth of each network and the number of neurons per hidden layer. 
We note that for the Deaths from Cancer and Flu vaccination rates data sets, we employ in-sample testing, which is due to their very small size of $n = 31$; see Table~\ref{tbl:datasets}.
We further note that the MNIST data set contains $70,\!000$ images of which $60,\!000$ are for training and $10,\!000$ for testing \cite{BALD19}. However, for our experiments we only made use of a balanced subset $1,\!000$ images, which we deemed sufficient to test our robustness model.

\begin{table}[htbp]
\centering
\small
\renewcommand{\arraystretch}{1.0}
\begin{tabularx}{\textwidth}{X c c X c c c X}
\toprule
Data set & Input width & \#HL & \#Neurons & AF & \#OU & AF & Train/Test \\
\midrule
Deaths from Cancer &
1 &
2 &
(64, 32) &
ReLU &
1 &
linear &
In-sample \\
Flu vaccination rates &
1 &
2 &
(64, 32) &
tanh &
1 &
linear &
In-sample \\
Heart disease &
28 &
2 &
(16, 8) &
ReLU &
2 &
sigmoid &
75/25 \\
Breast Cancer &
30 &
2 &
(32, 16) &
ReLU &
2 &
softmax &
75/25 \\
MNIST &
784 &
3 &
(512, 256, 128) &
ReLU &
10 &
softmax &
98.36/1.64 \\
\bottomrule
\end{tabularx}
\caption{Summary of neural network architectures used for each data set; \#HL denotes the number of hidden layers, AF denotes the activation function, \#OU denotes the number of output units, and Train/Test denotes the train and test split percentage.}
\label{tbl:arch}
\end{table}
\smallskip

We report in Table~\ref{tbl:base-regress} the baseline performance in terms of MSE and RMSE for the FNNs deployed for the proof-of concept regression tasks, shown in Table~\ref{tbl:datasets}.  Furthermore, in Table~\ref{tbl:base-class} we report the baseline performance in terms of the standard metrics of Micro F1 and Macro F1 \cite{LEVE25} (noting that Micro F1 is equal to accuracy \cite{OPTI24}), and MBS for the FNNs deployed for the classification tasks, shown in Table~\ref{tbl:datasets}. As further validation for the MNIST data set, we also computed the baseline performance on the full 10,000 image data set, and, as such, the differences to those shown in the third row of Table~\ref{tbl:base-class} were not statistically significant.

\begin{table}[!htp]
\centering
\small
\setlength{\tabcolsep}{8pt}
\renewcommand{\arraystretch}{1.15}
\begin{tabular}{l c c c}
\toprule
Data set              & MSE     & RMSE   & MAE \\
\midrule
Deaths from Cancer    & 10.1927 & 3.1926 & 1.9622 \\
Flu vaccination rates & 12.8175 & 3.5802 & 2.44   \\
\bottomrule
\end{tabular}
\caption{Baseline performance on the proof-of-concept regression data sets.}
\label{tbl:base-regress}
\end{table}

\begin{table}[!htp]
\centering
\small
\setlength{\tabcolsep}{8pt}
\renewcommand{\arraystretch}{1.15}
\begin{tabular}{l c c c c}
\toprule
Data set                & Micro F1 & Macro F1 & MBS \\
\midrule
Heart disease & 0.8533   & 0.8532   & 0.1059 \\
Breast Cancer & 0.972    & 0.9702   & 0.0203 \\
MNIST         & 0.979    & 0.979    & 0.0137 \\ 
\bottomrule
\end{tabular}
\caption{Baseline performance on the classification data sets.}
\label{tbl:base-class}
\end{table}
\smallskip 

For comparison in Table ~\ref{tbl:base-lit}, we report the performance, in terms of accuracy (recalling that accuracy is equal to Micro F1), from studies using the same classification data and FNN architectures. 
It can be seen that the baseline performance of our models is close enough to what has been reported in the literature. This is important, since robustness is not informative when the baseline performance is not adequate. Note that in Table~\ref{tbl:base-lit} we only compare to the classification data sets from Table~\ref{tbl:datasets}, since the two regression data sets are small data sets we use for proof-of-concept only.

\begin{table}[!htp]
\centering
\small
\setlength{\tabcolsep}{8pt}
\renewcommand{\arraystretch}{1.15}
\begin{tabular}{lclc}
\toprule
Data set & Approximate accuracy & Reference \\
\midrule
Heart disease &
0.8--0.86 &
\cite{ASGA24} \\
Breast Cancer &
0.86--0.97 &
\cite{SHAR18} \\
MNIST &
0.96--0.99 &
\cite{BALD19} \\
\bottomrule
\end{tabular}
\caption{Indicative performance ranges reported in the literature for comparable models.}
\label{tbl:base-lit}
\end{table}
\smallskip

We now let $\theta = \rho(\sigma, \alpha, {\rm NN})$ as computed by Algorithm~\ref{alg:robust}, with $\alpha = 0.05$ and NN being the FNN used, as described in Table~\ref{tbl:arch} for each data set. (The choice of $\alpha = 0.05$ is pragmatic in terms of our experimentation, although we are aware that different data sets may, in principle, require different values of $\alpha$, say $0.01$.) In the experiments $\rho(\sigma, \alpha, {\rm NN})$, has been tested on a range of $\sigma$ values starting from $0$ and increased by increments of $0.01$ until $\sigma = 1$; see \cite{ZUR09}.  When $m$, the number of numeric features, is greater than $1$, $p$, the perturbation probability, takes on the four values, $0.25,0.5,0.75$ and $1$, otherwise, when $m=1$, then $p=1$; see Subsection~\ref{subsec:real}. In the subsection that follows we demonstrate how we can summarise the numerical results using the concept of {\em robustness curves} (see \cite{SIRC21}).

\smallskip

\begin{figure}[!htp]
    \centering
    \includegraphics[scale=0.25]{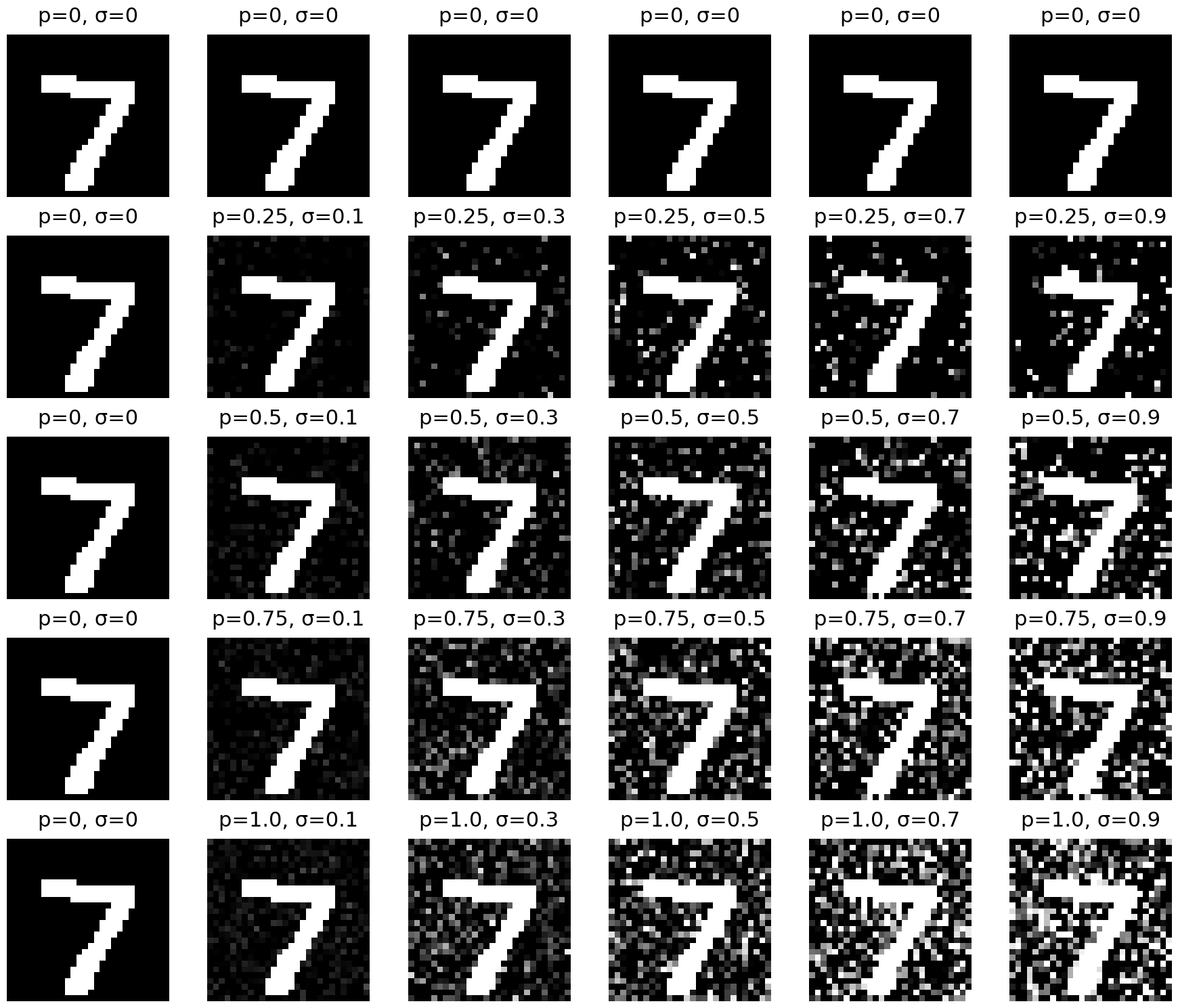}
    \caption{An example from the MNIST test data of several perturbations of the digit class $7$. (Note that when no perturbation occurs, we have $p=0$ and $\sigma = 0$.)}
    \label{fig:mnist_perturb}
\end{figure}

As an example of perturbed inputs from the MNIST test data, we show in Figure~\ref{fig:mnist_perturb} how the digit class $7$  is affected by increasing the perturbation. In the first column we have the unperturbed digit, while along the rest of the columns we have increasing values of $\sigma$: $0.1$, $0.3$, and $0.5$, with $p$ taking on the four values: $0.25,0.5,0.75$ and $1$ along the corresponding rows of the figure. As the value of these parameters increases, we observe how noise is gradually introduced to the sample.

\subsection{Robustness curves}
\label{subsec:curves}

A robustness curve is a plot of $\rho(\sigma, \alpha, {\rm NN})/{\rm MSE}_0$ (comprising the $y$-values of the plot) against $\sigma$ (comprising the $x$-values of the plot). That is $\rho(\sigma, \alpha, {\rm NN})$ is normalised by the baseline MSE, ${\rm MSE}_0$, where $\alpha$ and NN are understood from context as discussed towards the end of Subsection~\ref{subsec:numerics}. Thus, the theoretical origin point of a robustness curve, that is $(0,1)$, corresponds to $\rho(\sigma, \alpha, {\rm NN})/{\rm MSE}_0 =1$; in practice, as noted in Section~\ref{sec:robust}, for small $\sigma$, ${\rm MSE}\epsilon$ may fluctuate around ${\rm MSE}_0$ due to stochastic variation. When $y > 1$ the point $(x, y)$ on the curve gives the multiple of ${\rm MSE}_0$ corresponding to $\rho(\sigma, \alpha, {\rm NN})$, with $\sigma > 0$. 

\smallskip

Therefore, the robustness curve is scale-free, which will be useful for comparing the robustness of different data sets. So, for example if $y= \rho(\sigma, \alpha, {\rm NN})/{\rm MSE}_0 = 2$ then the robustness is twice the baseline MSE, ${\rm MSE}_0$.  For reference purposes, when using the RMSE as a measure matching the units of target values, the normalised robustness would be $\sqrt{2}$ times the baselines RMSE, where ${\rm RMSE}_0 = \sqrt{{\rm MSE}_0}$.

\smallskip
  
Now, for each data set we fit the robustness curves to the {\em Gompertz function} \cite{SEBE03,TJOR17}, which is widely used for growth processes, saturation effects, and cumulative phenomena. We note that the Gompertz curve is sigmoidal (that is, S-shaped) but unlike the logistic curve, it is asymmetric \cite{SEBE03}. We employ the standard form of he 4-parameter Gompertz function \cite{TJOR17}, given by
\begin{equation} \label{eq:gompertz}
f(x) = d + (a - d)\,\exp\!\left( -\exp\!\left( -b(x - c) \right) \right),
\end{equation}
where $exp(\cdot)$ is the exponential function, having the 4 parameters,
\begin{itemize}
    \item $a$ is the {\em upper asymptote} (maximum value of $y$),
    \item $d$ is the {\em lower asymptote} (minimum value of $y$),
    \item $b$ is the {\em growth-rate parameter} controlling the steepness of the curve, and
    \item $c$ is the {\em inflection point}, i.e.\ the value of $x$ where the growth rate is maximal (horizontal shift of the curve).
\end{itemize}
\smallskip

For data sets with more than one numeric feature we have four robustness curves one for each value of $p$ (the perturb or not probability) and when there is only one numeric feature we have only one robustness curve corresponding to $p=1$. 

\smallskip

We measure the goodness-of-fit of the Gompertz curve with the $R^2$, {\em the coefficient of determination}, defined as,
\begin{equation}\label{eq:r2}
\frac{\sum_{i=1}^n \left({\rm NN}(x_i) - y_i \right)^2}{\sum_{i=1}^n \left(\bar{y} - y_i \right)^2},
\end{equation}
where $\bar{y}$ is the mean of the $y_i$'s, that is, $\bar{y} = \sum_{i=1}^n y_i/n$.
So, $R^2$ is the MSE divided by the MST ({\em mean total sum of squares}, which is essentially the sample variance of $y$), where the MST is the denominator of (\ref{eq:r2}) divided by $n$. Thus $R^2$ measures the 
proportion of explained variance.

\smallskip

On each robustness curve we indicate three significant data points:
\begin{enumerate}
\item The {\em inflection} point, defined above, which is the point of maximum growth.
\item The {\em knee} point, which is the maximum {\em curvature point} \cite{SEBE03,OSBO21} before inflection, observing that below the knee there is a linear regime. 
\item The {\em half-maximum slope} point, which is the intermediate threshold point between the knee and an inflection point, where the slope reaches half of its maximum.
\end{enumerate}
\smallskip

The qualitative properties of the Gompertz curve coincide with theoretical expectations for ${\rm MSE}\epsilon$ under increasing noise:
\begin{enumerate}[label=(\arabic*)]   
    \item {\em Flat initial regime}:
    For small~$\sigma$, perturbations lie within a locally linear region of the network. This corresponds to the early section of the Gompertz curve before the knee, where the MSE increases only slowly. The linear regime may be quantified by the first–order Taylor expansion of the Gompertz curve around the initial segment, where curvature is minimal.
    
    \item {\em Rapid degradation}:
    As $\sigma$ increases, increasing perturbations compounds the error. This matches the region of the Gompertz curve between the knee and the inflection point, where the slope is increasing. In this regime, the curvature of the Gompertz function grows rapidly, paralleling the nonlinear accumulation of prediction error.

    \item {\em Saturation}:
    When predictions become effectively noise-dominated, further increases in $\sigma$ cause only marginal changes in the prediction error. This corresponds to the region after the inflection point, where the Gompertz curve approaches its asymptote.  Here the slope decreases toward zero, reflecting the model losing any meaningful predictive power.
\end{enumerate}
\smallskip

In Figures \ref{fig:gompertz1} $\!-\!$ \ref{fig:gompertz5}, appearing in Appendix~\ref{app:gompertz}, we show the robustness curves for all the data sets. Correspondingly we show the fitted parameters for the curves in Tables \ref{tbl:gompertz1}, \ref{tbl:gompertz2}, \ref{tbl:gompertz3}, \ref{tbl:gompertz4} and \ref{tbl:gompertz5}, also appearing in Appendix~\ref{app:gompertz}. For all data sets the $R^2$ is very high, indicating very good fits to the Gompertz curve.

\smallskip

Recall that the $y$-values of the robustness curve are normalised robustness values for $x$-values representing $\sigma$, with $\sigma \le 1$. In Table~\ref{tbl:gompertz6} we show the $x$ and $y$-values, $x_{knee}$ and $y_{knee}$, of the $knee$, and also $x$-values capped by $y$-values, where the $y$-value is either equal to $2$ or to its value when it is less than $2$ at $x=1$, and denote these $x$ and $y$-values by $x_2$ and $y_2$, respectively. The $(x_{knee}$, $y_{knee})$-values indicate the threshold of initial flat regime, where the growth rate of $y$-values is linear, while the $(x_2, y_2)$-values indicate a reasonable user-defined threshold capped by at most twice the baseline MSE. From Table~\ref{tbl:gompertz6} we can conclude that for ascertaining robustness, for the regression data sets (Deaths from Cancer and Flu vaccination rates) using the $x_2$-values is the sensible strategy, while for the classification data sets (Heart disease, Breast Cancer and MNIST) using the $x_{knee}$ values is more sensible.

\begin{table}[!htp]
\centering
\begin{tabular}{llllll}
\toprule
Data set  & $p$ & $x_{knee}$ & $y_{knee}$ & $x_2$ & $y_2$ \\
\midrule
Deaths from Cancer    & 1    & 0.4080 & 17.6412 & 0.0546 & 2.0000 \\
Flu vaccination rates & 1    & 0.2604 & 6.7941  & 0.0727 & 2.0000 \\
Heart disease         & 0.25 & 0.1545 & 1.0248  & 1.0000 & 1.3674 \\
Heart disease         & 0.5  & 0.1607 & 1.0904  & 1.0000 & 1.5996 \\
Heart disease         & 0.75 & 0.1703 & 1.1414  & 1.0000 & 1.7837 \\
Heart disease         & 1    & 0.1607 & 1.1807  & 1.0000 & 1.9360 \\
Breast Cancer         & 0.25 & 0.0710 & 1.0848  & 0.6051 & 2.0000 \\
Breast Cancer         & 0.5  & 0.0346 & 1.0556  & 0.4332 & 2.0000 \\
Breast Cancer         & 0.75 & 0      & 0.9474  & 0.3513 & 2.0000 \\
Breast Cancer         & 1    & 0      & 0.8801  & 0.3030 & 2.0000 \\
MNIST                 & 0.25 & 0.3335 & 1.4806  & 0.4647 & 2.0000 \\
MNIST                 & 0.5  & 0.3042 & 2.2519  & 0.2818 & 2.0000 \\
MNIST                 & 0.75 & 0.2413 & 2.6755  & 0.2033 & 2.0000 \\
MNIST                 & 1    & 0.1872 & 2.8039  & 0.1527 & 2.0000 \\
\bottomrule
\end{tabular}
\caption{Thresholds on the $x$ and $y$-values of the Gompertz curve, which assist in interpreting the robustness of a NN model.}
\label{tbl:gompertz6}
\end{table}
\smallskip

To introduce the concept of a {\em robustness index}, which allows us to compare Gompertz robustness curves independently of scale, we remove the vertical parameters $a$ and $d$ so that the shape of the curve depends only on $b$ (the growth-rate) and $c$ (the inflection point). This is achieved by normalisation as follows: 
\begin{equation}\label{eq:gompertz-norm}
f_{norm}(x) = \frac{f(x)-d}{a-d} = \exp\!\left( -\exp\!\left( -b \left( x - c \right) \right) \right),
\end{equation}
where $f(x)$ is the Gompertz equation given in (\ref{eq:gompertz}). The interpretation of $b$ (growth-rate) and $c$ (inflection point) is that a large growth-rate implies steeper growth, while a smaller one implies a slower growth, and a smaller inflection point implies earlier rise, while a larger one implies a later rise. 

\smallskip

For two curves $f_{norm}^i$ and $f_{norm}^j$ with growth-rate and inflection point parameters $b_i, c_i$ and $b_j, c_j$, respectively, define the raw crossing point
\begin{equation}\label{eq:cross}
x^*_{ij} = \frac{b_i c_i - b_j c_j}{b_i - b_j}, \quad b_i \neq b_j,
\end{equation}
which is the solution $x^*_{ij}$ to $f_{norm}^i(x) = f_{norm}^j(x)$ when $b_i \neq b_j$.

\smallskip

To account for the relative growth rates of the curves, define the {\em effective crossing fraction} $x_\text{eff}$ as
\begin{equation}\label{eq:x_eff}
x_\text{eff} =
\begin{cases}
1 - x^*_{ij}, & \text{if } b_i < b_j,\\[1mm]
x^*_{ij}, & \text{if } b_i > b_j.
\end{cases}
\end{equation}
\smallskip

This ensures that the robustness correctly reflects the fraction of the interval over which $f_{norm}^i(x) \le f_{norm}^j(x)$. The {\em relative robustness index} (or simply the robustness index) of curve $i$ relative to curve $j$ over the interval $[0,1]$, denoted by $R_{ij}$, is then defined as
\begin{equation}\label{eq:Rij}
R_{ij} = \max(0, \min(1, x_\text{eff})).
\end{equation}
\smallskip

$R_{ij}$ measures the fraction of the interval $[0,1]$ over which $f_{norm}^i(x)\le f_{norm}^j(x)$ and captures all scenarios:
\begin{itemize}
    \item $R_{ij}=1$: curve $i$ is entirely below curve $j$ (maximal robustness),
    \item $R_{ij}=0$: curve $i$ is entirely above curve $j$ (no robustness),
    \item $0<R_{ij}<1$: curve $i$ crosses curve $j$, partial robustness over a subinterval of $[0,1]$.
\end{itemize}
\smallskip

In the special case when $b_i = b_j$, the curves differ only by a horizontal shift determined by the inflection points. In this case, since the Gompertz curve is strictly increasing, $R_{ij}$ is determined entirely by $c_i$ and $c_j$, as follows:
\begin{equation}\label{eq:Rij_equal_b}
R_{ij} =
\begin{cases}
1, & \text{if } c_i > c_j,\\[2mm]
0, & \text{if } c_i < c_j,\\[2mm]
1, & \text{if } c_i = c_j \ (\text{by convention}).
\end{cases}
\end{equation}
\smallskip

We can then define {\em overall robustness index}, denoted by $G_i$, as 
\begin{equation}\label{eq:curve-dom}
G_i = \frac{1}{\kappa-1} \sum_{j \neq i} R_{ij},
\end{equation}
where $\kappa$ is the total number of curves being analysed. 

\smallskip

This index is normalised to the unit interval $[0,1]$ and captures both full, over $[0,1]$, and partial robustness over a subinterval of $[0,1]$. $G_i = 1$ indicates that curve $i$ is more robust than all other curves over the full interval $[0,1]$, while $0 < G_i < 1$ reflects the average robustness over only part of the domain.  

\smallskip

In Table~\ref{tbl:robustness_pairs} we present the robustness relationships between the curves for the five data sets and demonstrate this with the fitted data, and in Table~\ref{tbl:robustness_dataset} we present the overall robustness index for the data sets indicating which $i$ is assigned to which data set.

\begin{table}[!htp]
\centering
\begin{tabular}{lllllll}
\toprule
$i$ & $j$ & $b_i$ & $c_i$ & $b_j$ & $c_j$ & $R_{ij}$ \\
\midrule
1 & 2 & 1.7122 & 0.9701 & 2.042  & 0.7317 & 1 \\
1 & 3 & 1.7122 & 0.9701 & 2.8214 & 0.5018 & 1 \\
1 & 4 & 1.7122 & 0.9701 & 1.2998 & 0.7045 & 1 \\
1 & 5 & 1.7122 & 0.9701 & 4.541  & 0.3991 & 0.9471 \\
2 & 3 & 2.042  & 0.7317 & 2.8214 & 0.5018 & 1 \\
2 & 4 & 2.042  & 0.7317 & 1.2998 & 0.7045 & 0.78 \\
2 & 5 & 2.0420 & 0.7317 & 4.541  & 0.3991 & 0.873 \\
3 & 4 & 2.8214 & 0.5018 & 1.2998 & 0.7045 & 0.329 \\
3 & 5 & 2.8214 & 0.5018 & 4.541  & 0.3991 & 0.7742 \\
4 & 2 & 1.2998 & 0.7045 & 2.0420 & 0.7317 & 0.22 \\
4 & 3 & 1.2998 & 0.7045 & 2.8214 & 0.5018 & 0.671 \\
4 & 5 & 1.2998 & 0.7045 & 4.5410 & 0.3991 & 0.723 \\
5 & 1 & 4.541  & 0.3991 & 1.7122 & 0.9701 & 0.053 \\
5 & 2 & 4.541  & 0.3991 & 2.0420 & 0.7317 & 0.127 \\
5 & 3 & 4.541  & 0.3991 & 2.8214 & 0.5018 & 0.226 \\
5 & 4 & 4.541  & 0.3991 & 1.2998 & 0.7045 & 0.277 \\
\bottomrule
\end{tabular}
\caption{Robustness relationships between normalized Gompertz curves using parameters with 4 decimal points precision. All pairs with $R_{ij}>0$ are included. Rows are ordered so that curve $i$ is more robust (lies below) relative to curve $j$. The robustness index, $R_{ij}$, measures the fraction of the interval $[0,1]$ over which $f_{norm}^i(x) \le f_{norm}^j(x)$.}
\label{tbl:robustness_pairs}
\end{table}

\begin{table}[!htp]
\centering
\begin{tabular}{lll}
\toprule
Data set              & $i$ & $G_i$ \\
\midrule
Deaths from Cancer    & 1  & 0.9866 \\
Flu Vaccination Rates & 2  & 0.6633 \\
Heart Disease         & 3  & 0.2758 \\
Breast Cancer         & 4  & 0.4035 \\
MNIST                 & 5  & 0.1708 \\
\bottomrule
\end{tabular}
\caption{Overall robustness index $G_i$ computed from the full $R_{ij}$ table. $G_i$ measures the average fraction of the interval $[0,1]$ over which curve $i$ is more robust (lies below) relative to the other curves. Higher $G_i$ indicates greater robustness.}
\label{tbl:robustness_dataset}
\end{table}
\smallskip

A clear hierarchy of the curves' robustness relative to each other emerges from Tables \ref{tbl:robustness_pairs} and \ref{tbl:robustness_dataset}. 
It can be seen that Deaths from Cancer is the most robust, since the curves for the other data sets are completely above it, apart from MNIST curve which almost fully above it. 
Second, Flu vaccination rates, which is more robust than Heart Disease and to a large degree more robust than Breast Cancer and MNIST. Third, Heart disease is to a large degree more robust than MNIST but less robust than Breast Cancer, while Breast Cancer is to a large degree more robust than both Heart disease and MNIST. Finally, MNIST is the least robust relative to the other data sets.
Then, where $>$ means {\em more robust than}, the following hierarchy arises in the interval $[0,1]$ between the data sets,
\begin{equation}\label{eq:hierarchy}
1 > 2 > 4 > 3 > 5.
\end{equation} 

\section{Concluding remarks}
\label{sec:conc}

We have presented a method for quantifying the robustness of a trained neural network model when its inputs are perturbed with Gaussian noise. Given a trained neural network, NN and standard deviation $\sigma$, $\rho(\sigma, \alpha, {\rm NN})$, defined in (\ref{eq:robust-mse}), gives us the robustness of the NN, with Gauassian perturbation, ${\cal N}(0,\sigma^2)$, and significance level, $\alpha$. The robustness, $\rho(\sigma, \alpha, {\rm NN})$, is then approximated with Algorithm~\ref{alg:robust}.

\smallskip

We have then described detailed experimentation using five real-world data sets trained with FNNs, showing how the robustness method we have presented can be put into practice.

\smallskip

In Table~\ref{tbl:gompertz6} we showed that using a user-defined threshold of twice the baseline MSE for the regression data sets (Deaths from Cancer and Flu vaccination rates), while for the classification data sets (Heart disease, Breast Cancer and MNIST) setting the threshold at the knee point will assist in interpreting the robustness of the NN model. 
Furthermore, building on Tables \ref{tbl:robustness_pairs} and \ref{tbl:robustness_dataset} using robustness curves we were able to show in (\ref{eq:hierarchy}) the following hierarchy between the data sets, 
\begin{displaymath}
{\rm Death \ from \ Cancer} > {\rm Flu \ vaccination \ rates} > {\rm Breast \ Cancer} > {\rm Heart \ disease} > {\rm MNIST}.
\end{displaymath} 

It is important to validate the method presented herein on additional real-world data sets with a variety of machine learning models.
Finally, an area of ongoing research is the extension of our notion of robustness to {\em variational autoencoders} (VAEs) \cite{KING19}, which are a type of generative neural network model, which include a probabilistic latent space. 

\appendix

\section{Figures of the Gompertz curves and tables of fitted parameters for the data sets}
\label{app:gompertz}

\begin{figure}[!htp]
    \centering
    \includegraphics[width=0.8\linewidth]{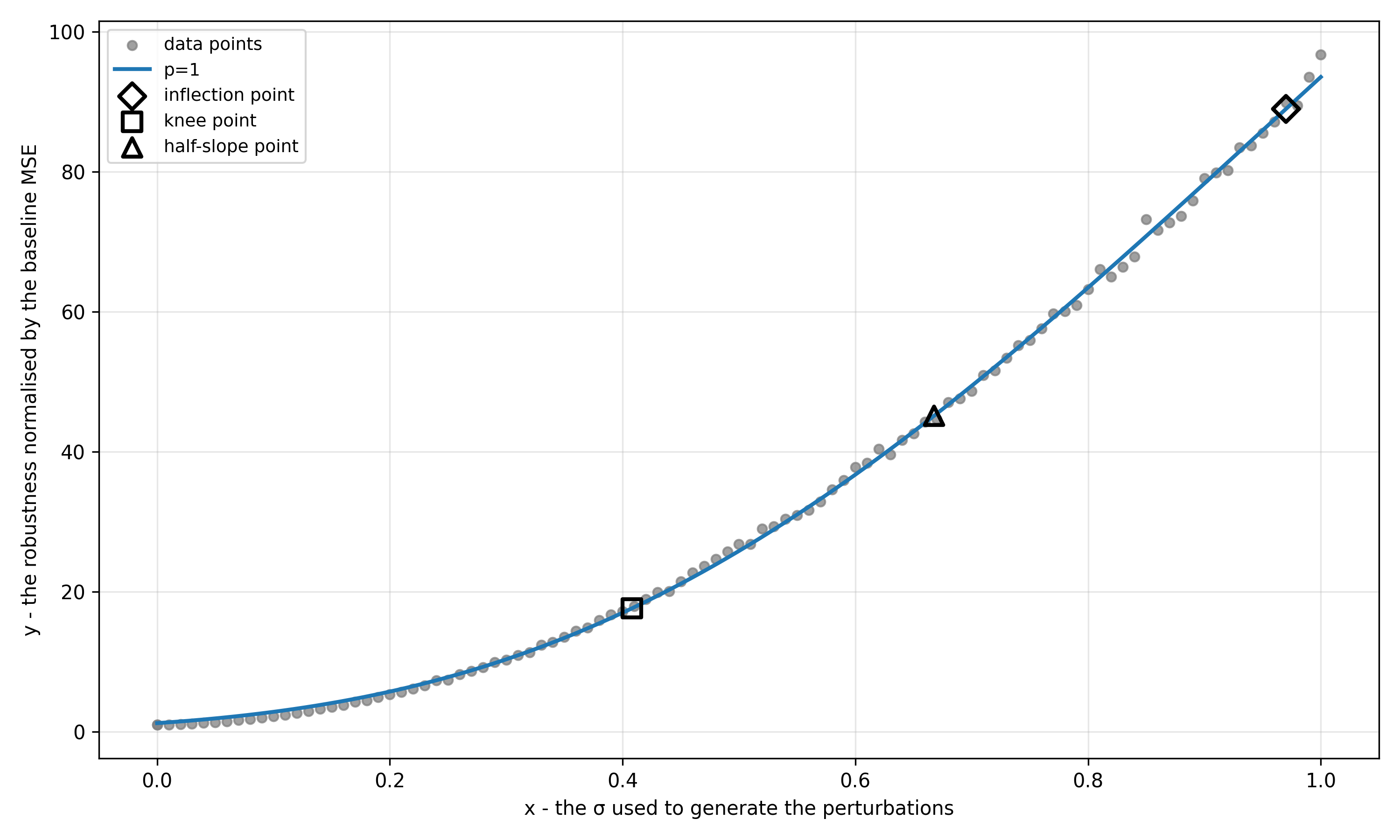}
    \caption{Gompertz fit of the robustness curve for the Deaths from Cancer data set.}
    \label{fig:gompertz1}
\end{figure}

\begin{table}[!htp]
\centering
\begin{tabular}{llllllll}
\toprule
$p$ & $a$      & $b$    & $c$    & $d$                   & $half \ slope$ & $knee$ & $R^2$ \\
\midrule
1   & 241.8389 & 1.7122 & 0.9701 & $6.137 \times 10^{17}$ & 0.6676        & 0.4079 & 0.9993 \\
\bottomrule
\end{tabular}
\caption{Gompertz fitted parameters for the robustness curve of the Deaths from Cancer data set. Parameters $a,b,c,d$ correspond to the four-parameter Gompertz model.}
\label{tbl:gompertz1}
\end{table}

\begin{figure}[!htp]
    \centering
    \includegraphics[width=0.8\linewidth]{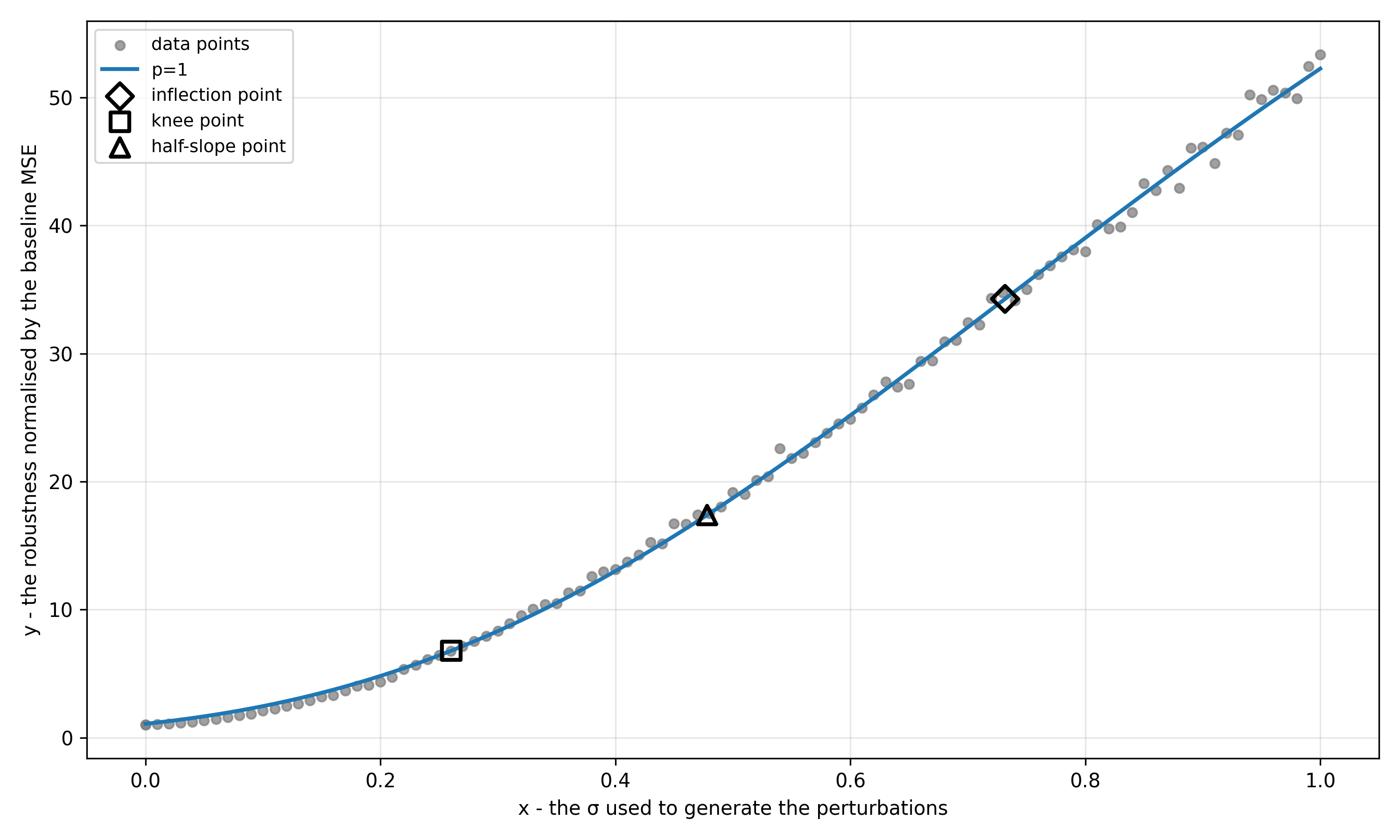}
    \caption{Gompertz fit of the robustness curve for the Flu vaccination rates data set.}
    \label{fig:gompertz2}
\end{figure}

\begin{table}[!htp]
\centering
\begin{tabular}{llllllll}
\toprule
$p$ & $a$     & $b$    & $c$    & $d$                    & $half \ slope$  & $knee$ & $R^2$ \\
\midrule
1   & 93.1385 & 2.042  & 0.7317 & $1.183 \times 10^{16}$ & 0.4781          & 0.2604 & 0.9988 \\
\bottomrule
\end{tabular}
\caption{Gompertz fitted parameters for the robustness curve of the Flu vaccination rates data set. Parameters $a,b,c,d$ correspond to the four-parameter Gompertz model.}
\label{tbl:gompertz2}
\end{table}

\begin{figure}[!htp]
    \centering
    \includegraphics[width=0.8\linewidth]{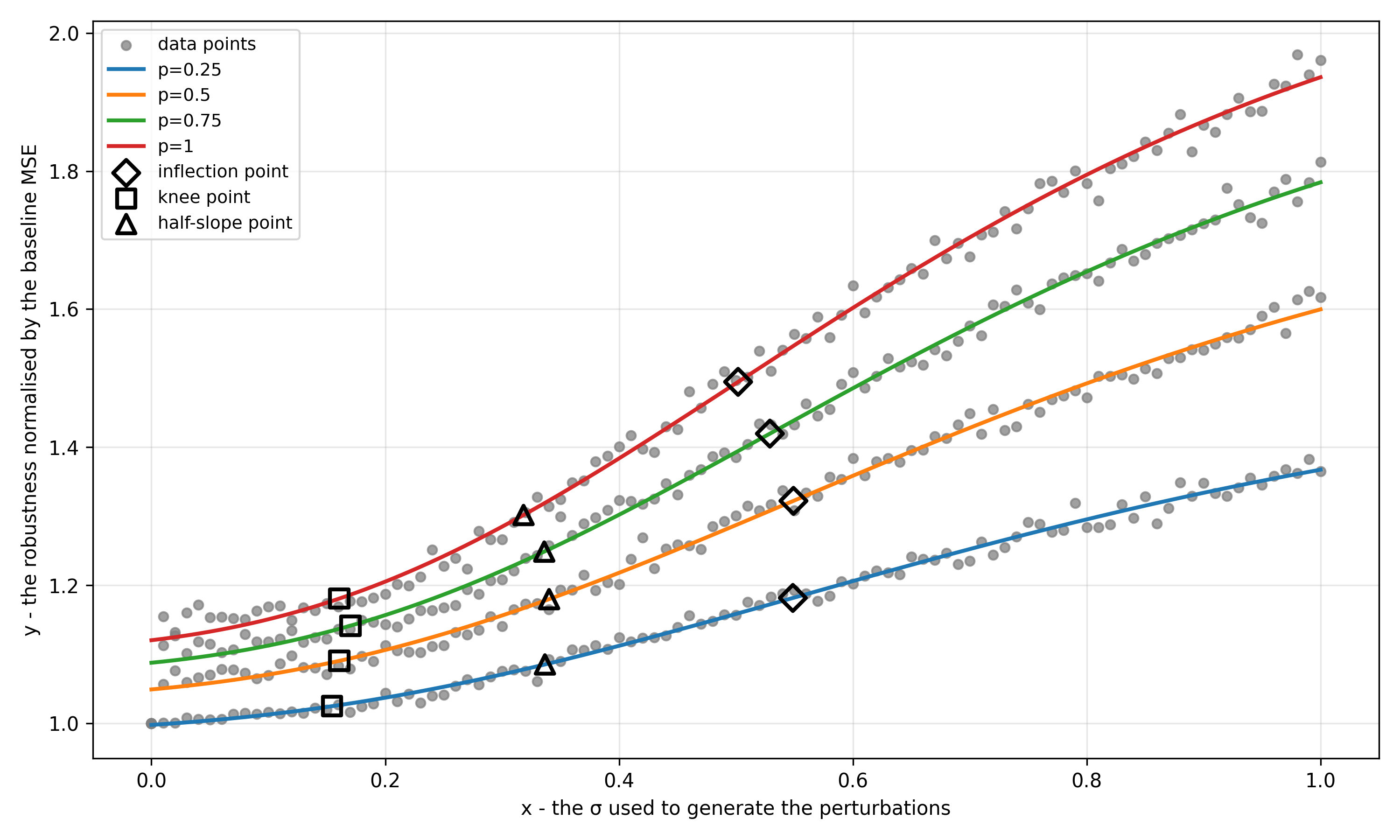}
    \caption{Gompertz fit of the robustness curve for the Heart disease data set.}
    \label{fig:gompertz3}
\end{figure}

\begin{table}[!htp]
\centering
\begin{tabular}{llllllll}
\toprule
$p$  & $a$    & $b$    & $c$    & $d$    & $half \ slope$  & $knee$ & $R^2$ \\
\midrule
0.25 & 1.5177 & 2.4418 & 0.5486 & 0.986  & 0.3366          & 0.1545 & 0.9932 \\
0.5  & 1.819  & 2.4774 & 0.5492 & 1.0331 & 0.3401          & 0.1607 & 0.9939 \\
0.75 & 2.0159 & 2.6829 & 0.529  & 1.0726 & 0.336           & 0.1703 & 0.9945 \\
1    & 2.1674 & 2.8214 & 0.5018 & 1.1031 & 0.3183          & 0.1607 & 0.9938 \\
\bottomrule
\end{tabular}
\caption{Gompertz fitted parameters for the robustness curves of the Heart disease data set. Parameters $a,b,c,d$ correspond to the four-parameter Gompertz model.}
\label{tbl:gompertz3}
\end{table}

\begin{figure}[!htp]
    \centering
    \includegraphics[width=0.8\linewidth]{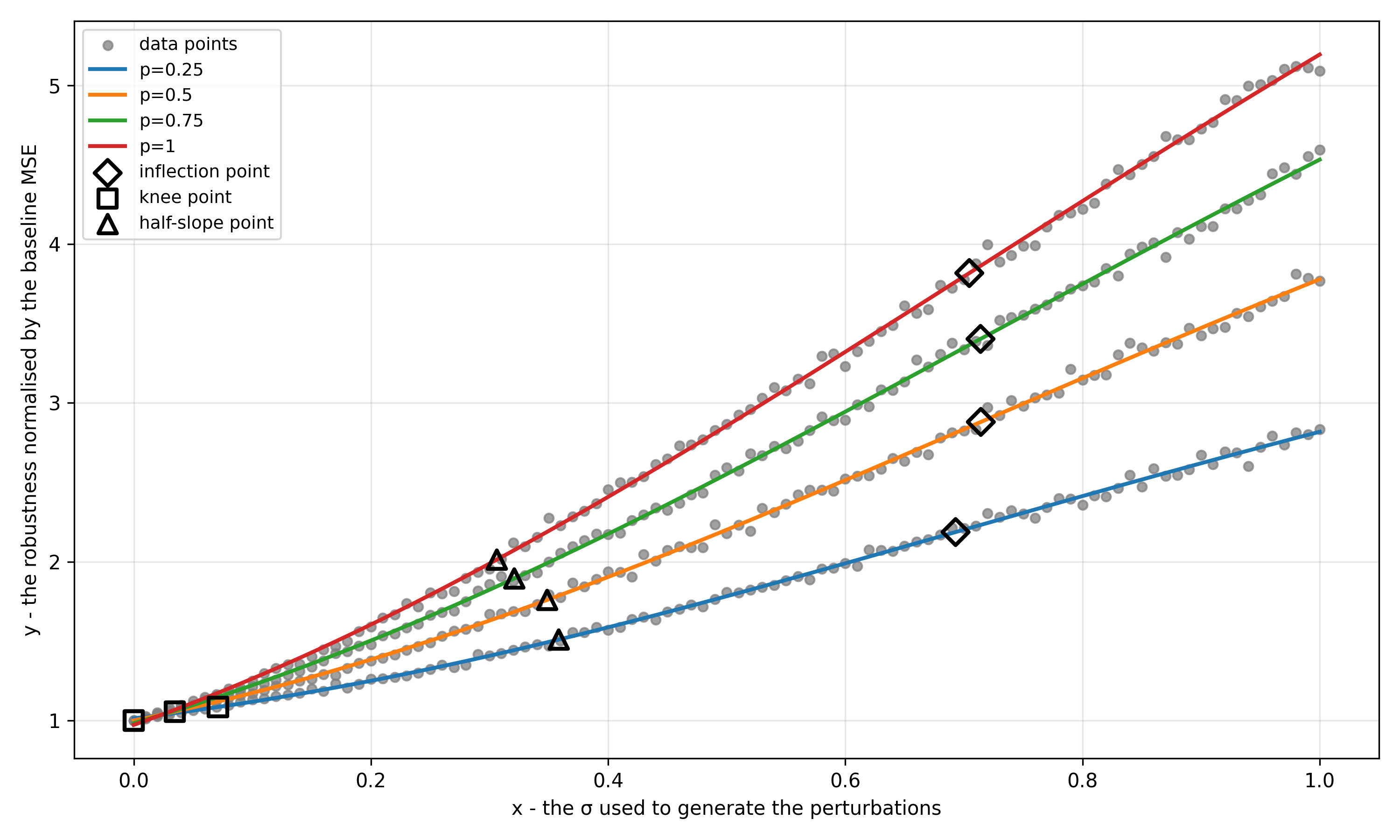}
    \caption{Gompertz fit for the robustness curves of the Breast Cancer data set.}
    \label{fig:gompertz4}
\end{figure}

\begin{table}[!htp]
\centering
\begin{tabular}{llllllll}
\toprule
$p$  & $a$     & $b$    & $c$     & $d$    & $half \ slope$ & $knee$  & $R^2$ \\
\midrule
0.25 & 4.5475  & 1.5474 & 0.693   & 0.8123 & 0.3584         & 0.071   & 0.9981 \\
0.5  & 6.7899  & 1.4159 & 0.7143  & 0.6044 & 0.3486         & 0.0345  & 0.9986 \\
0.75 & 8.6648  & 1.3171 & 0.714   & 0.3401 & 0.3209         & 0       & 0.999  \\
1    & 10.1129 & 1.2998 & 0.7045  & 0.1536 & 0.3062         & 0       & 0.9991 \\
\bottomrule
\end{tabular}
\caption{Gompertz fitted parameters for the robustness curves of the Breast Cancer data set. Parameters $a,b,c,d$ correspond to the four-parameter Gompertz model.}
\label{tbl:gompertz4}
\end{table}

\begin{figure}[!htp]
    \centering
    \includegraphics[width=0.8\linewidth]{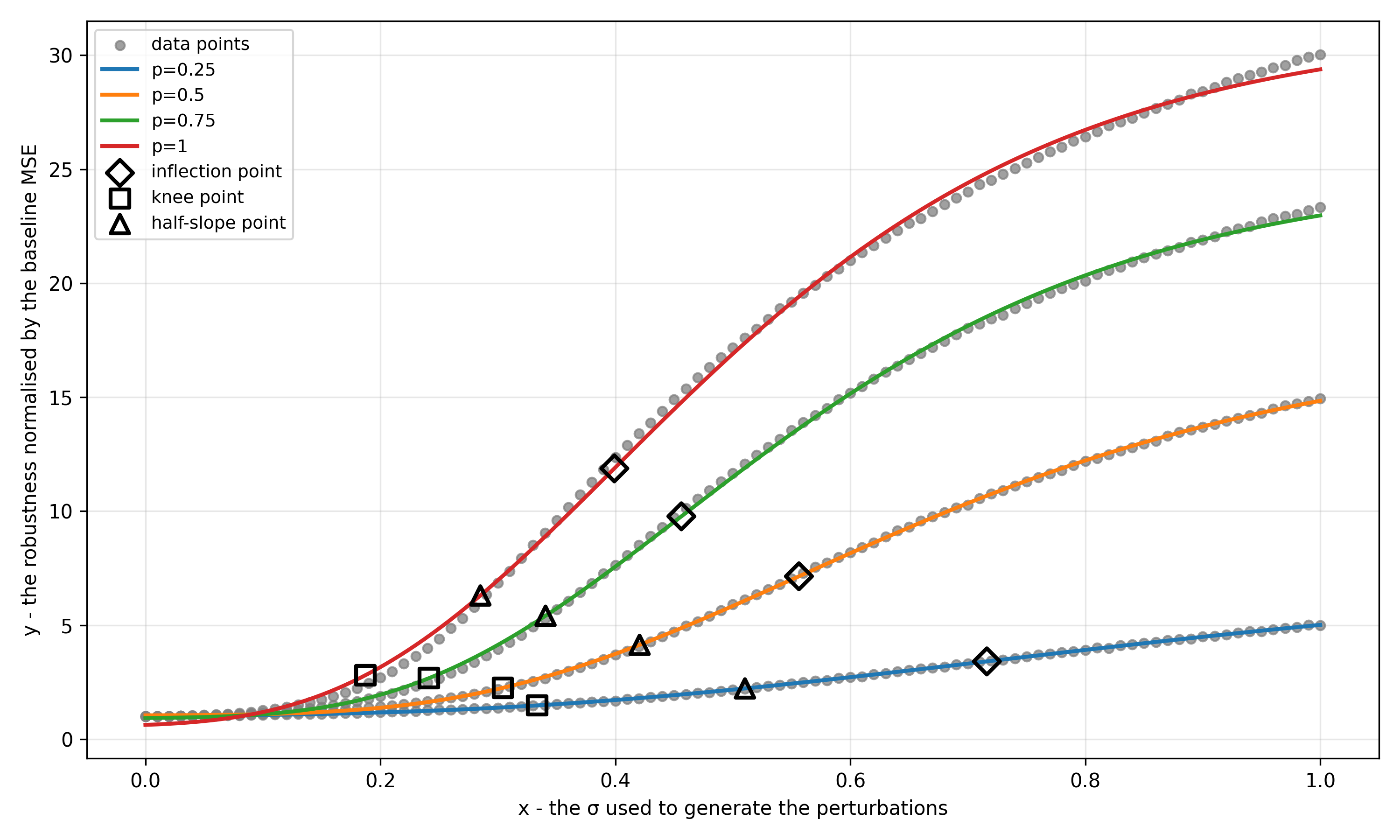}
    \caption{Gompertz fit of the robustness curves for the MNIST data set.}
    \label{fig:gompertz5}
\end{figure}

\begin{table}[!htp]
\centering
\begin{tabular}{llllllll}
\toprule
$p$  & $a$     & $b$    & $c$     & $d$    & $half \ slope$ & $knee$ & $R^2$ \\
\midrule
0.25 & 7.5501  & 2.5148 & 0.7162  & 1.003  & 0.5103         & 0.3335 & 0.9999 \\
0.5  & 17.6289 & 3.8178 & 0.5563  & 1.0419 & 0.4206         & 0.3042 & 0.9999 \\
0.75 & 24.9933 & 4.4781 & 0.4562  & 0.9194 & 0.3406         & 0.2413 & 0.9997 \\
1    & 31.3285 & 4.5410 & 0.3991  & 0.5594 & 0.2851         & 0.1872 & 0.9991 \\
\bottomrule
\end{tabular}
\caption{Gompertz fitted parameters for the robustness curves of the MNIST data set. 
Parameters $a,b,c,d$ correspond to the four-parameter Gompertz model.}
\label{tbl:gompertz5}
\end{table}

\newcommand{\etalchar}[1]{$^{#1}$}


\begin{thebibliography}{{OEC}26b}

\bibitem[Agg18]{AGGA18}
C.C. Aggarawal.
\newblock {\em Neural Networks and Deep Learning: A Textbook}.
\newblock Springer International Publishing, Cham, Switzerland, 2018.

\bibitem[ARV09]{ALIP09}
C.~Alippi, M.~Roveri, and G.~Vanini.
\newblock Robustness in neural networks.
\newblock In M.~{Khosrow-Pour}, editor, {\em Encyclopedia of Information
  Science and Technology}, pages 3314--3321. IGI Global, Hershey, PA, second
  edition, 2009.

\bibitem[Asg24]{ASGA24}
E.~Asgarov.
\newblock A comprehensive analysis of machine learning techniques for heart
  disease prediction.
\newblock {\em Open Access Library Journal}, 11:1--17, 2024.

\bibitem[BBB{\etalchar{+}}25]{BALE25}
A.~Balendran, C.~Beji, F.~Bouvier, O.~Khalifa, et~al.
\newblock A scoping review of robustness concepts for machine learning in
  healthcare.
\newblock {\em npj Digital Medicine}, 8(Article number 38):9 pages, 2025.

\bibitem[BSI19]{BALD19}
A.~Baldominos, Y.~Saez, and P.~Isasi.
\newblock A survey of handwritten character recognition with {MNIST} and
  {EMNIST}.
\newblock {\em Applied Sciences}, 9:Article 3169, 16 pages, 2019.

\bibitem[CD14]{CHAI14}
T.~Chai and R.R. Draxler.
\newblock Root mean square error ({RMSE}) or mean absolute error ({MAE})? --
  {A}rguments against avoiding {RMSE} in the literature.
\newblock {\em Geoscientific Model Development}, 7:1247--1250, 2014.

\bibitem[Cle98]{UCHD26}
Cleveland.
\newblock {Heart Disease}.
\newblock \url{https://archive.ics.uci.edu/dataset/45/heart+disease}, 1998.
\newblock UCI Machine Learning Repository.

\bibitem[CWJ21]{CHIC21a}
D.~Chicco, M.J. Warrens, and G.~Jurman.
\newblock The coefficient of determination {R}-squared is more informative than
  {SMAPE}, {MAE}, {MAPE}, {MSE} and {RMSE} in regression analysis evaluation.
\newblock {\em PeerJ Computer Science}, 7:e623, 2021.

\bibitem[HB21]{HUI21}
L.~Hui and M.~Belkin.
\newblock Evaluation of neural architectures trained with square loss vs
  cross-entropy in classification tasks.
\newblock {\em Machine Learning Archive}, arXiv:2006.07322 [cs.LG], 2021.

\bibitem[HCC19]{HUST19}
T.~Huster, {C.-Y}.J. Chiang, and R.~Chadha.
\newblock Limitations of the {L}ipschitz constant as a defense against
  adversarial examples.
\newblock In {\em Proceedings of ECML PKDD 2018 Workshops, European Conference
  on Machine Learnin Principles and Practice of Knowledge Discovery in
  Databases (ECML PKDD)}, pages 16--29, Dublin, 2019.

\bibitem[Hoe26]{HOES26}
L.~Hoessly.
\newblock On misconceptions about the {B}rier score in binary prediction
  models.
\newblock {\em Global Epidemiology}, 11:Article 100242, 8 pages, 2026.

\bibitem[Ial19]{IALO19}
C.~Ialongo.
\newblock Confidence interval for quantiles and percentiles.
\newblock {\em Biochemia Medica}, 29:5--17, 2019.

\bibitem[JS11]{JAPK11}
N.~Japkowicz and M.~Shah.
\newblock {\em Evaluating learning algorithms: {A} classification perspective}.
\newblock Cambridge University Press, Cambridge, UK, 2011.

\bibitem[{Kag}19]{MNIS19}
{Kaggle Community}.
\newblock {MNIST} {D}ataset: {T}he {MNIST} database of handwritten digits.
\newblock \url{www.kaggle.com/datasets/hojjatk/mnist-dataset}, 2019.
\newblock Accessed via Kaggle; see original data set \cite{LECU98}.

\bibitem[KW19]{KING19}
D.P. Kingma and M.~Welling.
\newblock An introduction to variational autoencoders.
\newblock {\em Foundations and Trends in Information Retrieval}, 12:307--392,
  2019.

\bibitem[LBB98]{LECU98}
Y.~LeCun, L.~Bottou, and Y.~Bengio.
\newblock Gradient-based learning applied to document recognition.
\newblock {\em Proceeding of the IEEE}, 86:2278--2324, 1998.

\bibitem[LH25]{LEVE25}
M.~Levene and M.~Harris.
\newblock {\em Just Enough Data Science and Machine Learning: Essential Tools
  and Techniques}.
\newblock Addison-Wesley, Hoboken, NJ, 2025.

\bibitem[LL25]{LI25}
J.~Li and G.~Li.
\newblock Triangular trade-off between robustness, accuracy, and fairness in
  deep neural networks: {A} survey.
\newblock {\em ACM Computing Surveys}, 57:Article 140, 40 pages, 2025.

\bibitem[MMS{\etalchar{+}}18]{MADR18}
A.~Madry, A.~Makelov, L.~Schmidt, D.~Tsipras, et~al.
\newblock Towards deep learning models resistant to adversarial attacks.
\newblock In {\em Proceedings of the 6th International Conference on Learning
  Representations (ICLR)}, Vancouver, 2018.
\newblock 23 pages.

\bibitem[OCB21]{OSBO21}
J.M. Osborne, W.~Cook, and M.J. Boss\'{e}.
\newblock Quantifying the curvature of curves: an intuitive introduction to
  differential geometry.
\newblock {\em PRIMUS, Problems, Resources, and Issues in Mathematics
  Undergraduate Studies}, 31:133--152, 2021.

\bibitem[{OEC}26a]{OECD_deaths}
{OECD}.
\newblock {Deaths from Cancer in the USA from 1990 to 2020}.
\newblock \url{www.oecd.org/en/data/indicators/deaths-from-cancer.html}, 2026.

\bibitem[{OEC}26b]{OECD_flu}
{OECD}.
\newblock {Influenza Vaccination Rates in 2020 for 31 countries}.
\newblock
  \url{www.oecd.org/en/data/indicators/influenza-vaccination-rates.html}, 2026.

\bibitem[Opt24]{OPTI24}
J.~Optiz.
\newblock A closer look at classification evaluation metrics and a critical
  reflection of common evaluation practice.
\newblock {\em Transactions of the Association for Computational Linguistics},
  12:820--836, 2024.

\bibitem[PBVN24]{PRAK24}
A.~{Prakash R}, A.~Bhattacharyya, J.~Vaughan, and V.N. Nair.
\newblock Assessing robustness of machine learning models using covariate
  perturbations.
\newblock {\em Machine Learning Archive}, arXiv:1908.02729 [stat.ML], 2024.

\bibitem[RI99]{RAVI99}
Y.~Raviv and N.~Intrator.
\newblock Variance reduction via noise and bias constraints.
\newblock In A.J.C. Sharkey, editor, {\em Combining Artificial Neural Nets:
  Ensemble and Modular Multi-Net Systems}, Perspectives in Neural Computing,
  chapter~7, pages 163--178. Springer-Verlag, London, 1999.

\bibitem[Ros23]{ROSS23}
S.M. Ross.
\newblock {\em Simulation}.
\newblock Academic Press, London, UK, sixth edition, 2023.

\bibitem[SFMR23]{SCRU23}
L.~Scrucca, C.~Fraley, T.B. Murphy, and A.E. Raftery.
\newblock {\em Model-based clustering, classification, and density estimation
  using mclust in {R}}.
\newblock Chapman \& Hall/CRC, Boca Raton, FL, 2023.

\bibitem[SKD18]{SHAR18}
A.~Sharma, S.~Kulshrestha, and S.B. Daniel.
\newblock Machine learning approaches for cancer detection.
\newblock {\em International Journal of Engineering and Manufacturing},
  8:45--55, 2018.

\bibitem[{\v{S}}S21]{SIRC21}
J.~{\v{S}}ircelj and D.~Sko{\v{c}}aj.
\newblock Accuracy-perturbation curves for evaluation of adversarial attack and
  defence methods.
\newblock In {\em Proceedings of 25th International Conference on Pattern
  Recognition (ICPR)}, pages 6290--6297, Milan, Italy, 2021.

\bibitem[SW03]{SEBE03}
G.A.F. Seber and C.J. Wild.
\newblock {\em Nonlinear Regression}.
\newblock Wiley Series in Probability and Statistics. John Wiley {\&} Sons,
  Hoboken, NJ, 2003.

\bibitem[SZS{\etalchar{+}}14]{SZEG14}
C.~Szegedy, W.~Zaremba, I.~Sutskever, J.~Bruna, et~al.
\newblock Intriguing properties of neural networks.
\newblock {\em Computer Vision and Pattern Recognition Archive},
  arXiv:1312.6199v4 [cs.CV], 2014.

\bibitem[TT17]{TJOR17}
K.M.C. Tj{\o}rve and E.~Tj{\o}rve.
\newblock The use of {G}ompertz models in growth analyses, and new
  {G}ompertz-model approach: {A}n addition to the unified-{R}ichards family.
\newblock {\em {PLoS ONE}}, 12(6):e0178691, 2017.

\bibitem[Wis95]{WDBC26}
Diagnostic Wisconsin.
\newblock {Breast Cancer}.
\newblock
  \url{https://archive.ics.uci.edu/dataset/17/breast+cancer+wisconsin+diagnostic},
  1995.
\newblock UCI Machine Learning Repository.

\bibitem[ZJPD09]{ZUR09}
R.M. Zur, Y.~Jiang, L.L. Pesce, and K.~Drukker.
\newblock Noise injection for training artificial neural networks: A comparison
  with weight decay and early stopping.
\newblock {\em Medical Physics}, 36:4810--4818, 2009.

\bibitem[ZK25]{ZUHL25}
{M.-M.} Z\"uhlke and D.~Kundenko.
\newblock Adversarial robustness of neural networks from the perspective of
  {L}ipschitz calculus: {A} survey.
\newblock {\em ACM Computing Surveys}, 57:Article 142, 41 pages, 2025.

\end{thebibliography}
\end{document}